\documentclass{article}

\PassOptionsToPackage{numbers, compress}{natbib}

\usepackage[preprint]{neurips_2020}

\usepackage[utf8]{inputenc} 
\usepackage[T1]{fontenc}    
\usepackage{hyperref}       
\usepackage{url}            
\usepackage{booktabs}       
\usepackage{amsfonts}       
\usepackage{nicefrac}       
\usepackage{microtype}      

\PassOptionsToPackage{hyphens}{url}\usepackage{hyperref}

\usepackage{times}  
\usepackage{helvet}  
\usepackage{courier}  
\usepackage{mathtools}

\usepackage{xspace}
\usepackage{latexsym}
\usepackage{amsfonts}
\usepackage{amsmath,bm}
\usepackage{amssymb}
\usepackage{color}
\usepackage{colortbl}
\usepackage{pbox}
\usepackage{array,multirow,graphicx}
\usepackage{paralist}
\usepackage{enumerate}
\usepackage[color,matrix,arrow,all]{xy}
\usepackage{comment}
\usepackage{booktabs}
\usepackage{balance}
\usepackage{stmaryrd}
\usepackage{pifont}
\usepackage{hhline}
\usepackage{listings}
\usepackage{array}
\usepackage{float}
\usepackage[flushleft]{threeparttable}
\usepackage{graphicx,wrapfig,lipsum}
 \usepackage[framemethod=TikZ]{mdframed}
\usetikzlibrary{arrows}
\usepackage{rotating}

\DeclareMathAlphabet{\pazocal}{OMS}{zplm}{m}{n}

\usepackage{array}
\newcolumntype{M}{>{\begin{varwidth}{2cm}}l<{\end{varwidth}}}
\newcolumntype{L}[1]{>{\raggedright\let\newline\\\arraybackslash\hspace{0pt}}m{#1}}
\newcolumntype{C}[1]{>{\centering\let\newline\\\arraybackslash\hspace{0pt}}m{#1}}
\newcolumntype{R}[1]{>{\raggedleft\let\newline\\\arraybackslash\hspace{0pt}}m{#1}}

\usepackage{array}
\usepackage{url}
\usepackage{longtable}
\usepackage{multirow}
\usepackage{xcolor}
\usepackage{algorithmic}
\usepackage[ruled, noline, longend, linesnumbered, boxed]{algorithm2e}
\usepackage{mdwlist}
\usepackage{subcaption}

\title{Embarrassingly Simple MixUp for Time-series}

\author{Karan Aggarwal\thanks{This author’s work was completed prior to his
affiliation with Amazon.} \\
Department of Computer Science\\
University of Minnesota\\
Minneapolis, MN, USA \\
\texttt{aggar081@umn.edu} \\
\And
Jaideep Srivastava \\
Department of Computer Science\\
University of Minnesota\\
Minneapolis, MN, USA \\
\texttt{srivasta@umn.edu} \\
}

\begin{document}

\maketitle

\begin{abstract}
Labeling time series data is an expensive task because of domain expertise and dynamic nature of the data. Hence, we often have to deal with limited labeled data settings. Data augmentation techniques have been successfully deployed in domains like computer vision to exploit the use of existing labeled data. We adapt one of the most commonly used technique called MixUp, in the time series domain. Our proposed, MixUp++ and LatentMixUp++, use simple modifications to perform interpolation in raw time series and classification model’s latent space, respectively. We also extend these methods with semi-supervised learning to exploit unlabeled data. We observe significant improvements of 1\% - 15\% on time series classification on two public datasets, for both low labeled data as well as high labeled data regimes, with LatentMixUp++.
\end{abstract}

\section{Introduction}
Time series data is one of the most commonly found data in nature across various domains like healthcare, finance, astronomy, and meteorology. Labeling time series data is quite an intensive process owing to the temporal and dynamic nature of data, unlike other domains like images or text.

Time series data labeling usually requires domain expertise and time to gather labeled data in time series. This is particularly an acute problem in domains like healthcare where labeling has to be quite precise owing to high stakes. Hence, for many applications, particularly in healthcare, only limited labeled data is available.

Recently, there is an advent of machine learning models in time series analysis, in particular deep learning, which require large datasets for performing well. Also, training on small datasets suffers from issues like over-fitting~\cite{salman2019overfitting}. These issues can significantly impact the performance as well as generalizability of time series models.

Data Augmentation is a technique to augment a training dataset with artificially created examples. It has been used extensively in computer vision to expand the size of training image datasets~\cite{krizhevsky2012imagenet, shorten2019survey}. It involves either using data transformations or using generative models for creating artificial training samples. Commonly used data transformations on images are rotating the image, flipping, or modulating colors.

These data transformations are quite simple but specific to image domain, as such transformations can completely change the meaning of a time-series. This is compounded by the huge diversity in domains of time series data, with very limited generic data transformations. There have been limited attempts in time series domain to come up with such augmentations, mostly in the wearables domains~\cite{um2017data, steven2018feature}. They have found permutation or rotation based transformations to be most effective.

Deep generative models like Generative Adversarial Networks~\cite{goodfellow2014generative} are used to generate images from noise distribution or conditional on desired style~\cite{jing2019neural} or labels~\cite{mirza2014conditional}. These have recently been used for time series data augmentation as well~\cite{esteban2017real, ramponi2018t}. However, these methods are quite challenging to train and to generate real world images.

MixUp~\cite{zhang2017mixup} was proposed in computer vision, as a training method to interpolate two real data points to generate a synthetic sample. Model is trained on the interpolated data, and not on the original training data. However, it is specific to the image domain as two images can be superimposed over each other, but a linear interpolation of two time series can be quite different. It has been adapted to other domains like natural language processing interpolation of sentences in latent space for text classification~\cite{guo2019augmenting}.

In this work, we adapt MixUp by simple extensions to time series classification problems; we train the model with the interpolated data as well as original data, in a true data augmentation fashion. Next, we also perform multiple MixUp steps for each batch of original data. We perform the same operation in latent space, called LatentMixUp. Our proposed methods, MixUp++ and LatentMixUp++ outperform baselines: non-augmentation based as well as permutation based time series augmentation. The proposed methods, especially LatentMixUp works particularly well in a low labeled data regime. We further extend MixUp to semi-supervised settings with pseudo-labeling based MixUp. These methods do not need any additional effort or hyper-parameter tuning apart from classifier tuning while improving performance considerably.

In summary, we make following contributions:
\begin{itemize}
    \item We propose simple but effective extensions to MixUp, MixUp++ and LatentMixUp++ for time series classification;
    \item We propose Pseudo-labeling based MixUp for semi-supervised learning to leverage unlabeled data; and
    \item Our results on two datasets: human activity and sleep staging, shows that our methods outperform baselines considerably in both low labeled data regime as well as high labeled data setting.
\end{itemize}

We organize the rest of this work as follows. Section~\ref{sec4:related} discusses relevant literature in data augmentation. Section~\ref{sec4:methodology} discusses background concepts like MixUp and our proposed methods. Section~\ref{sec4:experiments} presents details on our experimental settings. We present our results in Section~\ref{sec4:results}. Finally, we summarize our conclusions in Section~\ref{sec4:conc}.

\section{Related Work}
In this section, we give an overview of data augmentation literature and data augmentation in time series. 

\subsection{Data Augmentation in Machine Learning}
Data augmentation is a technique of adding existing data with synthetically generated data. This synthetic generation of data can be done through: 1) Simple data transformations or, 2) using Generative machine learning methods. 

\paragraph{Simple Data Transformation Augmentations} These simple data transformations encode human knowledge about the data domain. For example, in computer vision tasks like classification, we know that a transformation like rotating an image or adjusting its colors would not change its class. These geometric image transformations like flipping, rotating, cropping, or noise injection have been extensively used in the computer vision domain to increase data set sizes~\cite{shorten2019survey}. Even the seminal ImageNet paper~\cite{krizhevsky2012imagenet}, used these transformations to increase the training dataset size. Taylor et al.~\cite{taylor2018improving} performed a empirical study on effect of different transformations on the data augmentation. AutoAugment~\cite{cubuk2018autoaugment} is a general augmentation technique for images based on reinforcement learning that learns appropriate augmentation from the given image data from a list of image augmentations. However, these methods are quite limited to images and do not consider characteristics like temporal nature in time series. In this work we explore MixUp~\cite{zhang2017mixup}, which interpolates two images for time-series data augmentation. We further explore MixUp in latent model embedding space as has been explored in domains like natural language processing with sentence mixup~\cite{guo2019augmenting}.

\paragraph{Generative Machine Learning Models} Generative machine learning models generate synthetic data from scratch by modeling the distribution of input data. The most prominent family of models are Generative Adversarial Networks (GANs)~\cite{goodfellow2014generative}. GANs can generate image data from a noise distribution by learning a function that transforms noise into an image by learning over real world images through adversarial learning.  Various methods in GANs have been proposed like Neural Style Transfer~\cite{jing2019neural}, which can create synthetic examples by converting a given input image with another theme like adding a hat to a person's headshot. Further, Conditional Adversarial Networks (CGANs)~\cite{mirza2014conditional}, have been proposed to generate images conditioned on a class label. Other generative methods include Variational Auto-Encoders (VAEs)~\cite{kingma2019introduction}. Generative models are however, quite difficult to train and to be used for generating realistic data.  

\subsection{Data Augmentation for Time Series Data}
There is limited work in the time series domain for data augmentation. Since time series data is so diverse and so domain specific, there are very limited generic geometric transformations in time series. Um et al.~\cite{um2017data} explored data transformations like rotation, scaling, permutation, or distance based dynamic time warping for data augmentation on wearables data. They found rotation and permutation transformations to be the most effective augmentations.  While permutation involves shuffling time series segments to create another sample, rotation involves 3D rotation of wearable time series. Steven Eyobu et al.~\cite{steven2018feature} also explored using scaling and jitter for wearable time series data augmentation. 

Generative models like GANs have been used for time series data augmentation like Esteban et al.~\cite{esteban2017real}. They generate medical time series like EEG using GANs with a recurrent neural network (RNN) backbone for both generator and discriminator to model the time series sequence. They showed that synthetic time series performed as well as real world time series data. Ramponi et al.~\cite{ramponi2018t} propose Time-Conditional Generative Adversarial Networks (T-CGAN), and use it to study the data augmentation for time series. They show considerable improvement on time series datasets by augmenting with GAN generated time series data.  However, these methods are challenging to train and tune to produce desirable results. 

In this work, we explore MixUp for time series data augmentation. We compare our method with permutation augmentation proposed by Um et al.~\cite{um2017data} and show the effectiveness of our simple method.  

\label{sec4:related}

\section{Methodology}
\label{sec4:methodology}
In this section, we present the methodology used in this work. We first give a background on MixUp technique, and next discuss our methodology.

\subsection{Supervised Setting: MixUp++ and LatentMixUp++}
\subsubsection{Mixup}
MixUp is a simple data augmentation technique proposed by Zheng et al.~\cite{zhang2017mixup} used in computer vision. It is based on a simple idea of a linear interpolation of two labeled training samples to produce a synthetic training sample. This results in: 1) increased number of labeled training samples; and 2) regularization in the supervised model's embedding space. It has shown encouraging results by improving the accuracy of supervised algorithms on image classification datasets. 

\noindent Given two training samples, ($x_1$, $y_1$) and ($x_2$, $y_2$), synthetic interpolated example can be given by:
\begin{eqnarray}
    \tilde{x} &=& \lambda x_1 + (1-\lambda) x_2\\
    \tilde{y} &=& \lambda y_1 + (1-\lambda) y_2
\end{eqnarray}

where, $\tilde{x}$ and $\tilde{y}$ are generated by example and label, respectively. $\lambda$ is ratio of mixing, which is sampled from a beta distribution as follows with a parameter $\alpha$:

\begin{eqnarray}
\lambda &\sim& \mathrm{Beta}(\alpha, \alpha)
\end{eqnarray}

\subsubsection{LatentMixUp}
MixUp performs an interpolation in the raw input space to create a new example. However, while it is intuitive to do such interpolation in the image domain, it is not as intuitive in domains in natural language or even time series. For example, interpolation of a cosine function and its opposite phase cosine, can produce a constant value time series, which might be unlikely to be observed. Hence, recent methods have proposed to perform mixup in latent space of classification model in domains like Natural language with SenMixUp~\cite{guo2019augmenting} where two sentences can not be interpolated directly. We refer to it as \emph{LatentMixUp}, to perform a mixup of two time series in the latent space, described as follows.

\noindent Let $f(x)$ be the classifier with $f: \mathbb{R}^T \longrightarrow \mathbb{R}^{|C|}$, where $C$ is the set of output classes. We can further define $f(x) = g(h(x))$, where $h(x)$ is an intermediate neural network representation. $g(x)$ is the remainder of neural network layers that are built on top of $h(x)$ to perform the classification. For example, in a 6 layered neural network, $h(x)$ can be the first 5 layers of the network, and $g(x)$ is the last layer of the neural network that performs the softmax for the classification task.

Given two training samples, ($x_1$, $y_1$) and ($x_2$, $y_2$), synthetic interpolated example can be given by:
\begin{eqnarray}
    h(\tilde{x}) &=& \lambda h(x_1) + (1-\lambda) h(x_2)\\
    \tilde{y} &=& \lambda y_1 + (1-\lambda) y_2
\end{eqnarray}

Final prediction on interpolated $h(\tilde{x})$ is done by passing it through $g(.)$. Hence, prediction for the newly interpolated example, $\tilde{y}_{pred} = g(h(\tilde{x}))$.
The key idea behind LatentMixUp is that it provides interpolation in the model's latent space which is arguably more linear than raw feature space, making interpolation operation more representative of the data manifold.

\subsubsection{Proposed: MixUp++ and LatentMixUp++}
In MixUp and LatentMixUp, the model is trained \emph{only} on the interpolated data, while the original data is not used. We hypothesize that this could impact the model performance as original training data is real world data and is important for creation of data manifold on which decision boundary is created. Further, MixUp is done by randomly permuting a training batch and interpolating it with the training batch examples in the given batch, with a sampled value of $\lambda$. While with enough epochs of training, the model could get to see enough pairs from all possible pairs with a good proportion of mixing coefficient $\lambda$, it is not as efficient as the number of epochs is usually limited. 

\noindent We propose two embarrassingly simple but effective additions to the MixUp training:
\begin{itemize}
\item \textbf{Train with original data}: We keep the original data during training instead of discarding the original data in MixUp~\cite{zhang2017mixup}.
\item \textbf{Train with multiple MixUps for a single data batch}: We perform mixup, $k \in \mathbb{N}$ times for a single training data batch, each with different sampled values of mixing coefficient $\lambda$ to increase the possible mixup examples model is trained on.
\end{itemize}

We hypothesize that training entirely on interpolated examples is not optimal for time series data, as interpolated examples can represent an entirely different phenomenon for time series. Hence, it is important to use the original real world data during training, and use the interpolated data as a regularizer.

\subsection{Semi-supervised Setting}
While MixUp is entirely based in a supervised learning setting, we further extend the notion of MixUp in the semi-supervised setting. For this purpose, we use pseudo-labeling, a simple but commonly used semi-supervised learning technique.

\subsubsection{Pseudo-Labeling}
Pseudo-labeling~\cite{lee2013pseudo} is a simple technique to leverage unlabeled data in presence of limited labeled data. The idea behind pseudo-labeling is simple: we train the model on the limited labeled data, perform inference on unlabeled data to choose the examples the model is quite confident on. These examples are then added to the training set with model predicted labels for the next training iteration. This process is continued until we the training stabilises or maximum number of epochs are reached. 

More formally, let labeled data be given as $\mathcal{D}_l=\{x_i^l, y_i^l\}_{i=1}^n$, where $n$ is the number of labeled samples. Let, $\mathcal{D}_u=\{x_j^u\}_{j=1}^m$ be an unlabeled dataset, with $m$ being the number of unlabeled samples. We train a classifier $f(x)$ with labeled and unlabeled examples. Firstly, labeled examples are used to train the model in the epoch, which is used to do inference over each unlabeled example $x_j^u$. If predicted $f(x^u) = \hat{y}^u \ge \tau$, then the pair $(x^u, y^u)$ is used to train the model further. $\tau$ is the confidence threshold to only select highly confident examples in training. $y^u \in \mathbb{R}^{|C|}$ is one hot encoding of highest probability class above the confidence threshold $\tau$ given by:

\begin{equation}
    y_c^u = 
        \begin{cases}
      1, & \text{if $c$ = $\mathrm{argmax}_{c \in |C|} \hat{y}^u$}\\
      0, & \text{otherwise}
    \end{cases}       
\end{equation}

Hence, we given a label of $y^u$ to the highest confident class of an unlabeled sample $x^u$, which is then used for training the model. Pseudo-label classification loss can be written as follows:

\begin{equation}
    \mathcal{L}(f, \mathcal{D}_l, \mathcal{D}_u) = \sum_{x^l \in \mathcal{D}_l} l(f, x^l, y^l) + \sum_{x^u \in \mathcal{D}_u} \mathbb{I}(f(x^u) \ge \tau) l(f, x^u, y^u)
\end{equation}

\noindent where, $l(f, x, y)$ is the standard cross-entropy classification loss. Note, value of threshold, $\tau$ is usually set to be high. We use a $\tau$ of 0.99 in this work.  A low value of $\tau$ would allow wrongly labeled samples from unlabeled data to find their way into the training data. This can destabilize the model, further reinforced by subsequent pseudo-labeling rounds.

\begin{figure}[t]
\includegraphics[width=\linewidth]{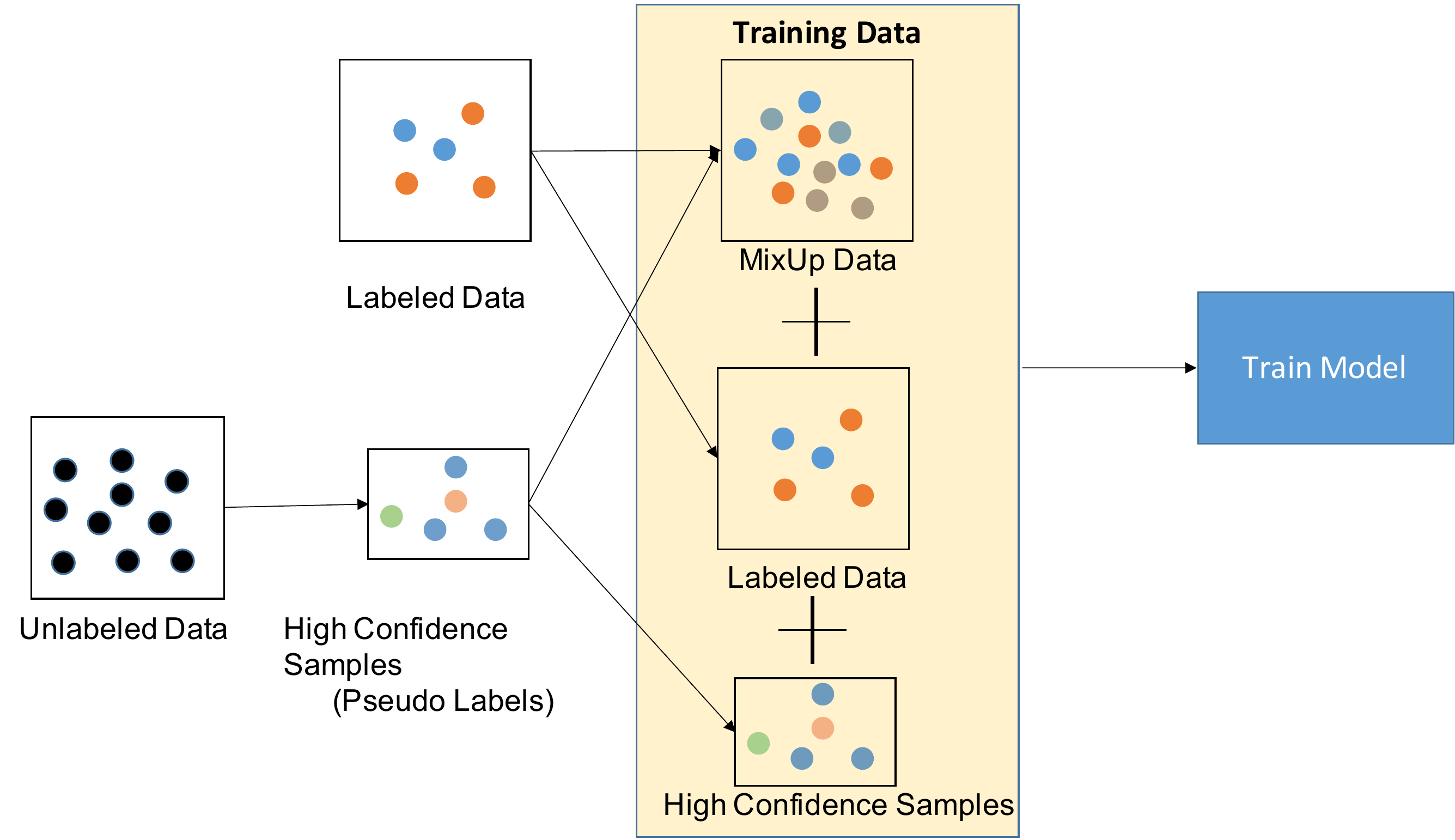}
\centering
\caption{ Schematic view of Pseudo-labeling with MixUp. First, we get the most confident samples from the unlabeled data, called pseudo-labels. We use these pseudo-labels and labeled data for MixUp. Finally, we train the model on three sources: labeled data, pseudo-labeled data, and  MixUp data.}
\centering
\label{fig:pseudo}
\end{figure}

\subsubsection{Pseudo-Labeling with MixUp}
We propose to leverage MixUp during pseudo-labeling in order to leverage the unlabeled data. The idea is simple: we add highly confident samples from unlabeled datasets with their pseudo-labels to the examples MixUp has access to. This way, MixUp interpolates between the labeled as well as pseudo-labeled samples from unlabeled data. Pseudo-Labeling with MixUp is shown in Figure~\ref{fig:pseudo}.

Next, we discuss our experimental settings to evaluate the effectiveness of our proposed methods: \emph{MixUp++} and \emph{LatentMixUp++}.

\section{Experimental Settings}
\label{sec4:experiments}
In this section, we present our experimental settings: datasets used, data augmentation baselines, and hyper-parameters tuned.

\subsection{Datasets}
\label{c4:datasets}
We use the following two datasets in this work related to human activity recognition and sleep staging:
\begin{itemize}
    \item \textbf{Sleep EDF}\footnote{\url{https://physionet.org/challenge/2012/}}: Sleep-EDF dataset~\cite{goldberger2000physiobank} contains polysomnography (PSG) for 20 subjects. We follow Supartak et al.~\cite{supratak2017deepsleepnet} by using the EEG channels (Fpz-Cz) sampled at 100 Hz, for sleep stage classification. Goal is to detect the sleep stage of each person for every 30 second window into Awake (W), deep sleep (N1, N2, N3), and Rapid Eye Movement (REM) sleep. This data has 25612 training and 8910 test samples. Note, the training and test sets are independent in terms of subjects following the implementation in DeepSleepNet\footnote{\url{https://github.com/akaraspt/deepsleepnet}}.
    \item \textbf{UCI HAR}\footnote{\url{https://archive.ics.uci.edu/ml/datasets/human+activity+recognition+using+smartphones}}: UCI HAR dataset~\cite{anguita2013public} contains multivariate time series across 6 channels of motion state recording of 30 volunteers, aged 19-48 years. It tracks 6 kinds of activities like walking, walking upstairs, walking downstairs, lying, sitting, or standing up. Data was collected using a waist bound Galaxy SII device with a sampling rate of 50 Hz. Task is to detect the state of the person for each 2.56 second window. This splitting gives about 7352 training set windows, and 2947 test set windows.
\end{itemize}

\subsection{Baselines Used}
We use the following data augmentation methods as a comparison with our proposed approach:
\begin{itemize}
    \item \textbf{Supervised Learning}: Baseline model without using any data augmentation, trained only on original real world labeled data.
    \item \textbf{PermuteAugment}: In this data augmentation technique, we permute segments of time series, based on Um \emph{et al.}~\cite{um2017data} where permutation and rotation were the most effective augmentation methods for time-series data. Since, rotation is specific to the dataset they used, we use permutation as the augmentation method as implemented in their code\footnote{\url{https://github.com/terryum/Data-Augmentation-For-Wearable-Sensor-Data}}.
    \item \textbf{MixUp}: We compare MixUp++ with the conventional MixUp algorithm.
    \item \textbf{LatentMixup}: This is just the conventional MixUp in latent space as described in Section~\ref{sec4:methodology}.  We implement this on the last layer before the softmax of our transformer neural network, described next.
\end{itemize}

For all these baselines, we use the same classifier, a transformer neural network~\cite{vaswani2017attention}. Using the same classifier ensures that we can confidently measure the difference between data augmentations of different methods.

\subsection{Hyper-parameters}
We use the same train-test split as provided in the datasets. For hyper-parameter tuning we randomly sample 80\% of the train set for training, and rest 20\% for validation. For the Sleep-EDF dataset, we use subject stratified split to ensure no overlap of subjects in training and validation sets. For transformer architecture, we tried the number of layers $\in \{1, 2, 3, 4, 5, 6\}$, and selected 5 layers and 5 as the number of heads. For layer size, we searched $\in \{32, 64, 100, 128, 160\}$ and selected 100 with a dropout of 0.15. We use Adam optimizer with a learning rate of $0.0002$. We searched for $\tau \in \{0.95, 0.97, 0.98,\\ 0.99, 0.995\}$ and selected 0.99 as the confidence threshold for pseudo labels.
Further, all experiments were performed 10 times with 10 different seeds to report the mean and standard deviations of all the results reported in the results section, Sec~\ref{sec4:results}.


\section{Results and Analysis}
\label{sec4:results}
In this section, we present results on the time series classification for two datasets: HAR and Sleep-EDF. Firstly, we present results in a fully supervised setting. Next, we show ablation results of performance as the proportion of training data. Finally, we show the performance of semi-supervised pseudo-labeling mixup.

\begin{table}
{
\caption{Time series classification results with various data augmentation methods measured on Accuracy/F1 micro, F1 Macro, and Cohen's Kappa for two datasets: HAR and Sleep-EDF. Note, differences between supervised and LatentMixUp++ are statistically significant with a $p < 0.01$ with student $t$-test.}
\label{table:mixupresults}
\begin{center}
\begin{tabular}{|l|ccc|} 
\toprule
\textbf{Method} & \textbf{Accuracy} (\%) & \textbf{F1 Macro} (\%) & \textbf{Kappa} (\%) \\
\midrule
\multicolumn{4}{|c|}{\textbf{\emph{HAR}}}\\
\midrule
Supervised   & 92.95 $\pm$ 0.83   & 92.99 $\pm$ 0.89   & 91.52 $\pm$ 1.0\\
\midrule
PermuteAugment    & 91.8 $\pm$ 0.91   & 91.97 $\pm$ 0.92     & 90.14 $\pm$ 1.09\\
PermuteAugment++   & 93.1 $\pm$ 0.43   & 93.26 $\pm$ 0.46     & 91.71 $\pm$ 0.51\\
\midrule
MixUp    & 92.63 $\pm$ 0.56   & 92.71 $\pm$ 0.65    & 91.14 $\pm$ 0.67\\
MixUp++   & 93.29 $\pm$ 0.80   & 93.38 $\pm$ 0.85   & 91.94 $\pm$ 0.97\\
MixUp++ (w/ 2 Iterations) & 93.45 $\pm$ 0.41   & 93.57 $\pm$ 0.45  & 92.13 $\pm$ 0.50\\
\midrule
LatentMixUp   & 94.07 $\pm$ 0.70   & 94.17 $\pm$ 0.73    & 92.87 $\pm$ 0.84\\
LatentMixUp++   & 94.41 $\pm$ 0.92   & 94.46 $\pm$ 0.95    & 93.28 $\pm$ 1.1\\
LatentMixUp++  (w/ 2 Iterations)  & \textcolor{blue}{\textbf{94.44}} $\pm$ 0.72   & \textcolor{blue}{\textbf{94.52}} $\pm$ 0.75   & \textcolor{blue}{\textbf{93.32}} $\pm$ 0.87\\
\midrule
\midrule
\multicolumn{4}{|c|}{\textbf{\emph{Sleep-EDF}}}\\
\midrule
\midrule
Supervised   & 80.57  $\pm$ 0.34  &  73.52  $\pm$ 0.85  & 73.48  $\pm$ 0.42\\
\midrule
PermuteAugment    & 74.21 $\pm$ 2.09 &   67.59 $\pm$ 2.31  &  64.3 $\pm$ 2.71\\
PermuteAugment++  &  78.89 $\pm$  0.35 &   71.75 $\pm$  0.52 & 71.63 $\pm$  0.43\\
\midrule
MixUp & 79.14 $\pm$ 0.96   &  66.3 $\pm$ 0.91  & 70.52 $\pm$ 1.47\\
MixUp++   & 80.47 $\pm$ 0.70  &  70.82 $\pm$ 1.62  & 72.8 $\pm$ 0.92\\
MixUp++  (w/ 2 Iterations) & 80.00 $\pm$ 0.57  &   68.7 $\pm$ 1.00 & 72.13  $\pm$ 0.67\\
\midrule
LatentMixUp   & 80.83  $\pm$ 0.82   & 72.71  $\pm$ 1.04  & 73.56  $\pm$ 1.09\\
LatentMixUp++     & 81.08  $\pm$ 0.56  &  73.74  $\pm$ 1.05  &  73.89  $\pm$ 0.65\\
LatentMixUp++ (w/ 2 Iterations)   & \textcolor{blue}{\textbf{81.12}}  $\pm$ 0.47  &  \textcolor{blue}{\textbf{73.79}}  $\pm$ 0.82 & \textcolor{blue}{\textbf{73.94}}  $\pm$ 0.62\\

\bottomrule
\end{tabular}
\end{center}
}
\end{table}

\begin{figure}
\includegraphics[width=0.5\linewidth]{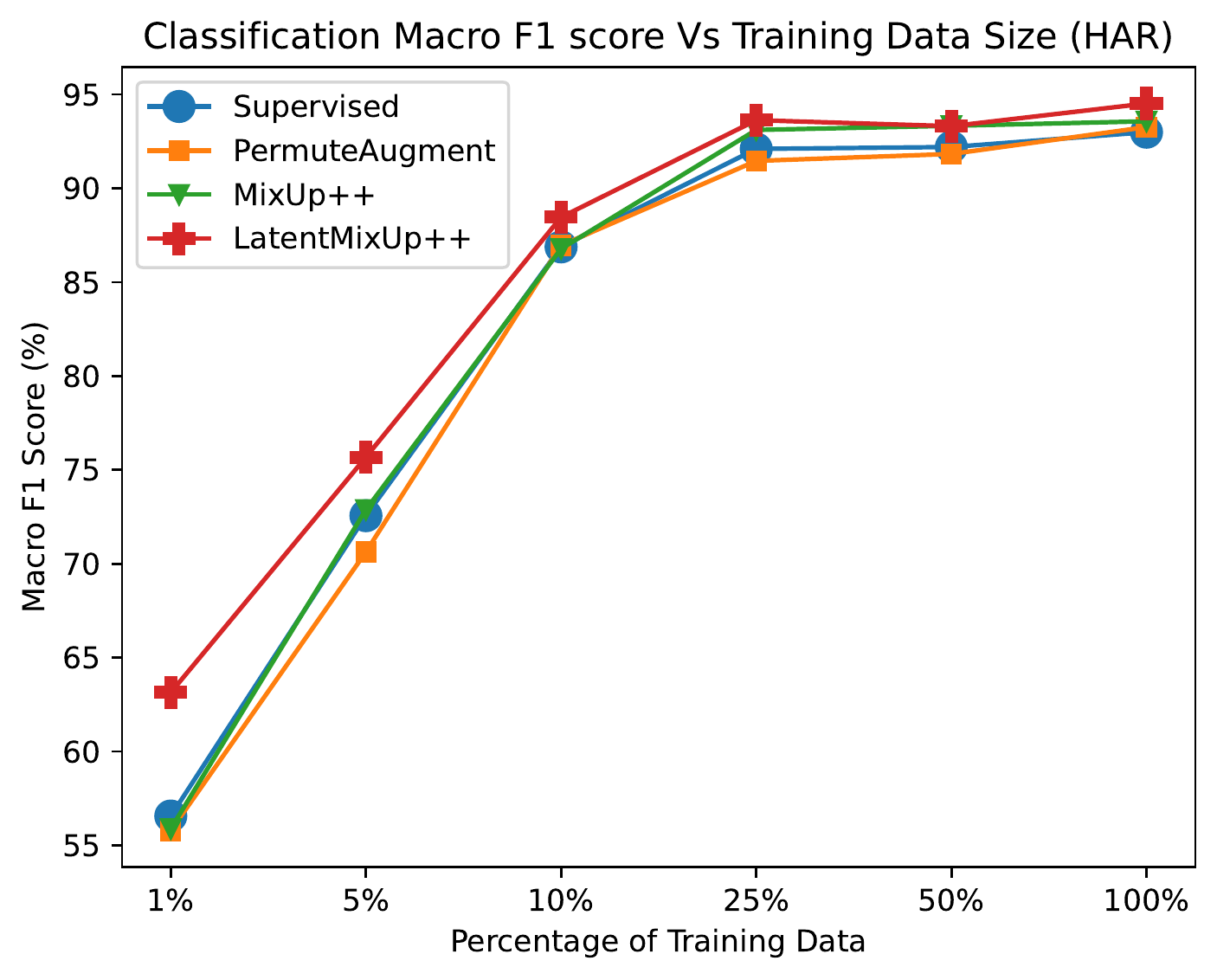}
\centering
\caption{ Classification Performance on HAR dataset as a function of percentage of training data. Note, MixUp++ and LatentMixUp++ performance is reported for models with 2 mixup batches.}
\centering
\label{fig:harAblation}
\end{figure}

\begin{figure}
\includegraphics[width=0.5\linewidth]{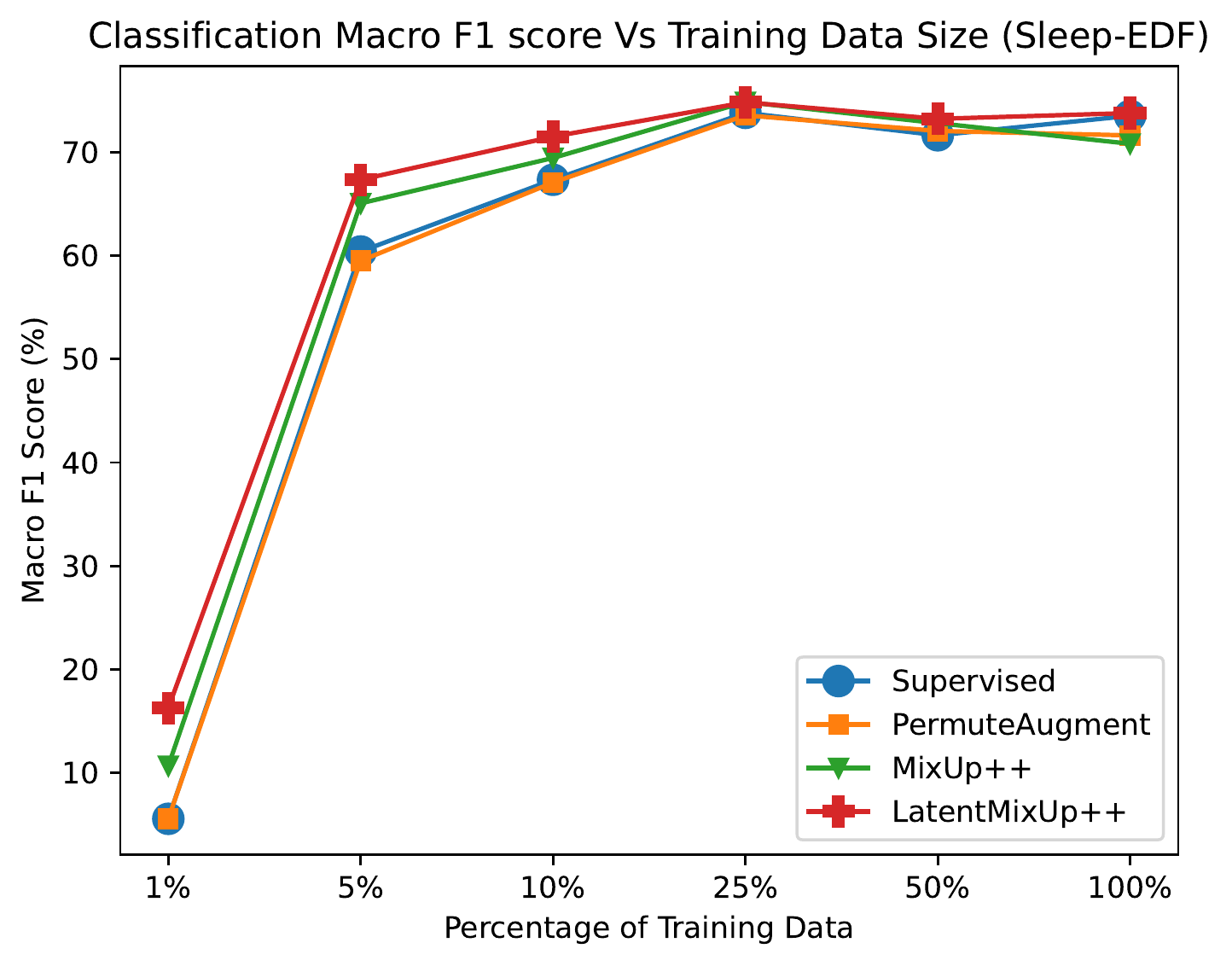}
\centering
\caption{ Classification Performance on Sleep-EDF dataset as a function of percentage of training data. Note, MixUp++ and LatentMixUp++ performance is reported for models with 2 mixup batches.}
\centering
\label{fig:sleepAblation}
\end{figure}

\subsection{Data Augmentation in Supervised Setting}
Classification results for the two datasets are presented in Table~\ref{table:mixupresults}. We observe that LatentMixUp with two batches of mixup, shows the best results, for the both datasets compared with the supervised baselines and other data augmentation baselines. Original MixUp~\cite{zhang2017mixup}, performs even lower than the vanilla supervised baseline while vanilla LatentMixUp performs only slightly better than the vanilla supervised baseline. The performance difference is especially stark in Sleep-EDF dataset. 

Adding batches of MixUp with original data helps MixUp, though all MixUp variants perform below or similar to supervised baseline. LatentMixUp standalone performs better than supervised baseline, while also benefiting significantly from the addition of original data in LatentMixUp++ variants. We can attribute this difference between MixUp and LatentMixUp to the fact that addition of two time series in raw time space could produce something that is completely unrecognizable, \textit{e.g.,} addition of two time out of sync consine functions could produce a zero-valued time series. Hence, mixing time series in the latent space with features more relevant to the time series class, is expected to produce better examples. Our hypothesis is confirmed by the empirical results.

\subsection{Ablation: Performance as a proportion of training Data}
We experiment to see the performance of our methods as a function of amount of training data, as shown in Figure~\ref{fig:harAblation} and Figure~\ref{fig:sleepAblation}. We report MixUp++ and LatentMixUp++ performance with 2 mixup batches. We observe that LatentMixUp++ still beats all the baseline methods for all the percentages of training data. \textbf{Additionally, LatentMixUp++ has a much higher F1 score differential low data regime (1\% or 5\%)}, than in high label data regime. This shows the effectiveness of LatentMixUp in regularizing the latent space for better classification performance. We note that MixUp++ is only marginally better than purely supervised method in low data regime for HAR dataset, but significantly better for Sleep-EDF dataset. We could attribute this to avoidance of over-fitting  by label regularization in low data regimes which are prone to over-fitting.

\begin{figure}
\includegraphics[width=0.5\linewidth]{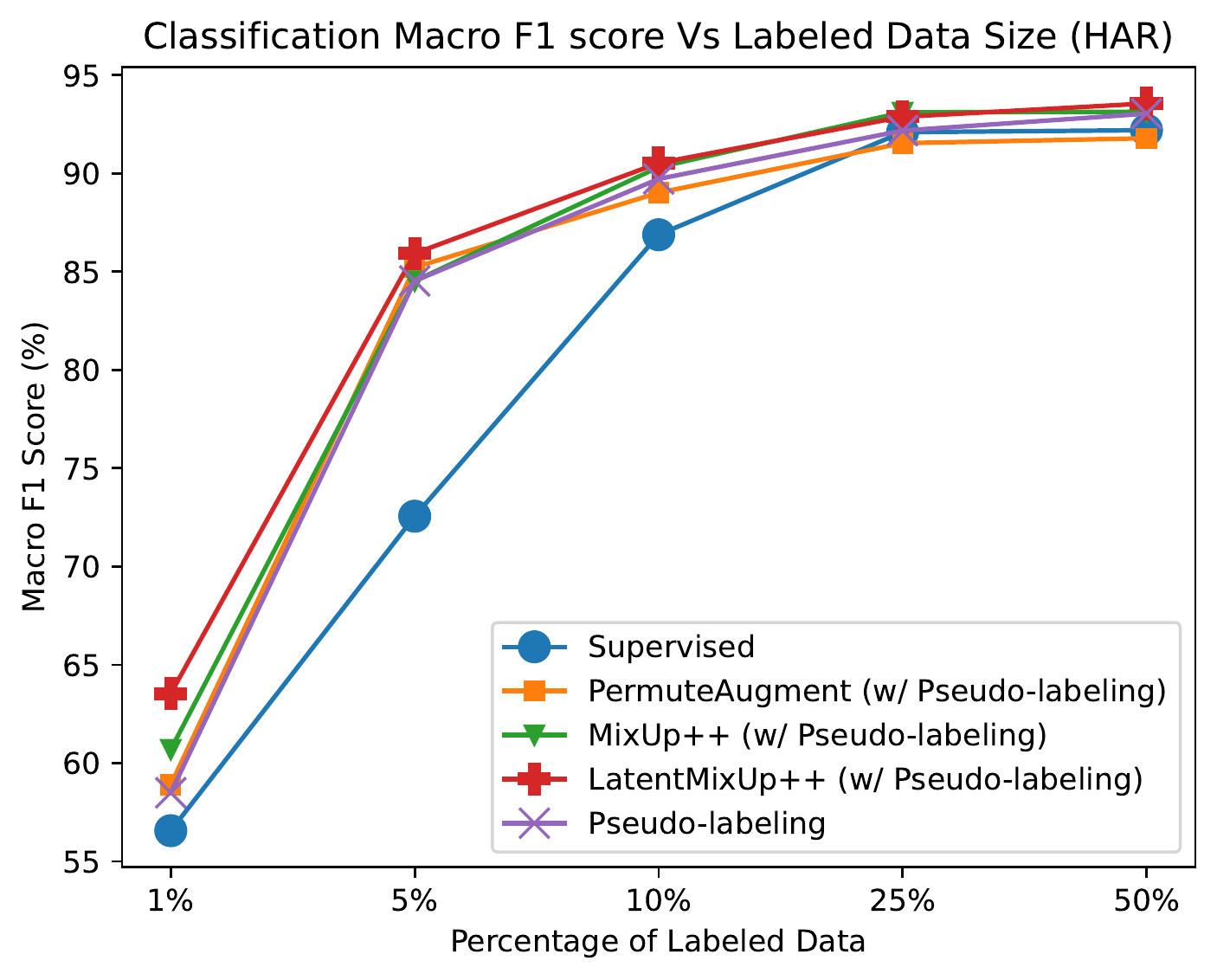}
\centering
\caption{Classification Performance on HAR dataset as a function of percentage of labeled data in the semi-supervised pseudo-labeling setting. Note, MixUp++ and LatentMixUp++ performance is reported for models with 2 mixup batches.}
\centering
\label{fig:harAblationPseudo}
\end{figure}

\begin{figure}
\includegraphics[width=0.5\linewidth]{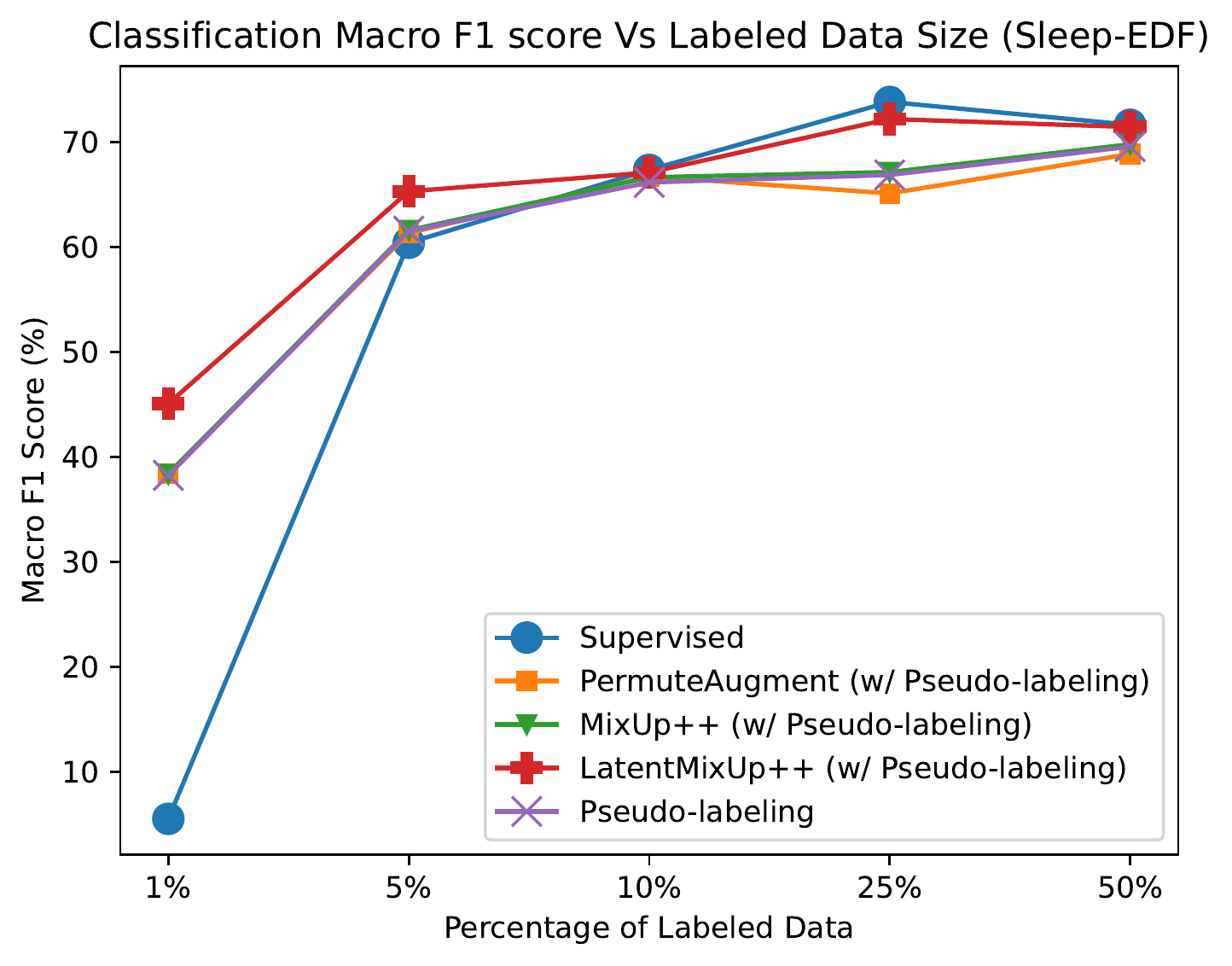}
\centering
\caption{ Classification Performance on Sleep-EDF dataset as a function of percentage of labeled data in the semi-supervised pseudo-labeling setting. Note, MixUp++ and LatentMixUp++ performance is reported for models with 2 mixup batches.}
\centering
\label{fig:sleepAblationPseudo}
\end{figure}

\subsection{Data Augmentation in Semi-Supervised Setting with Pseudo-labeling Mixup }
Next, we experiment with semi-supervised learning by using pseudo-labeling MixUp, as described in Section~\ref{sec4:methodology}. We used a confidence score of 0.99 to select highly confident examples for pseudo-labeling MixUp. Results are shown in Figure~\ref{fig:harAblationPseudo} and Figure~\ref{fig:sleepAblationPseudo} for HAR and sleep datasets, respectively. We report MixUp++ and LatentMixUp++ performance with 2 mixup batches. LatentMixUp++ still outperforms all other semi-supervised baselines, especially in the low labeled data regime with 6-7\% absolute increase in F1 score for the 1\% label data scenario versus traditional pseudo labeling for the two datasets. MixUp++ performs similar to  pseudo labeling baseline. The out-performance reduces to 0.5\% - 1\% in the higher labeled data regime like 50\%. 

Comparing performance of semi-supervised pseudo-labeling scenarios with purely supervised scenarios we see a drastic jump in low labeled data scenarios, especially for Sleep-EDF dataset. However, we do notice a drop in performance versus purely supervised method in the Sleep-EDF dataset as we increase labeled data to 25\%. This can be attributed to issues with pseudo-labeling as highly confident examples can bias the dataset, and propagate errors in initial training~\cite{zou2019confidence}. 

Based on these results, we can conclude that LatenMixUp++ with pseudo-labeling MixUp improves performance considerably in semi-supervised settings, especially for low labeled data regime.  

\begin{figure}
\includegraphics[width=0.5\linewidth]{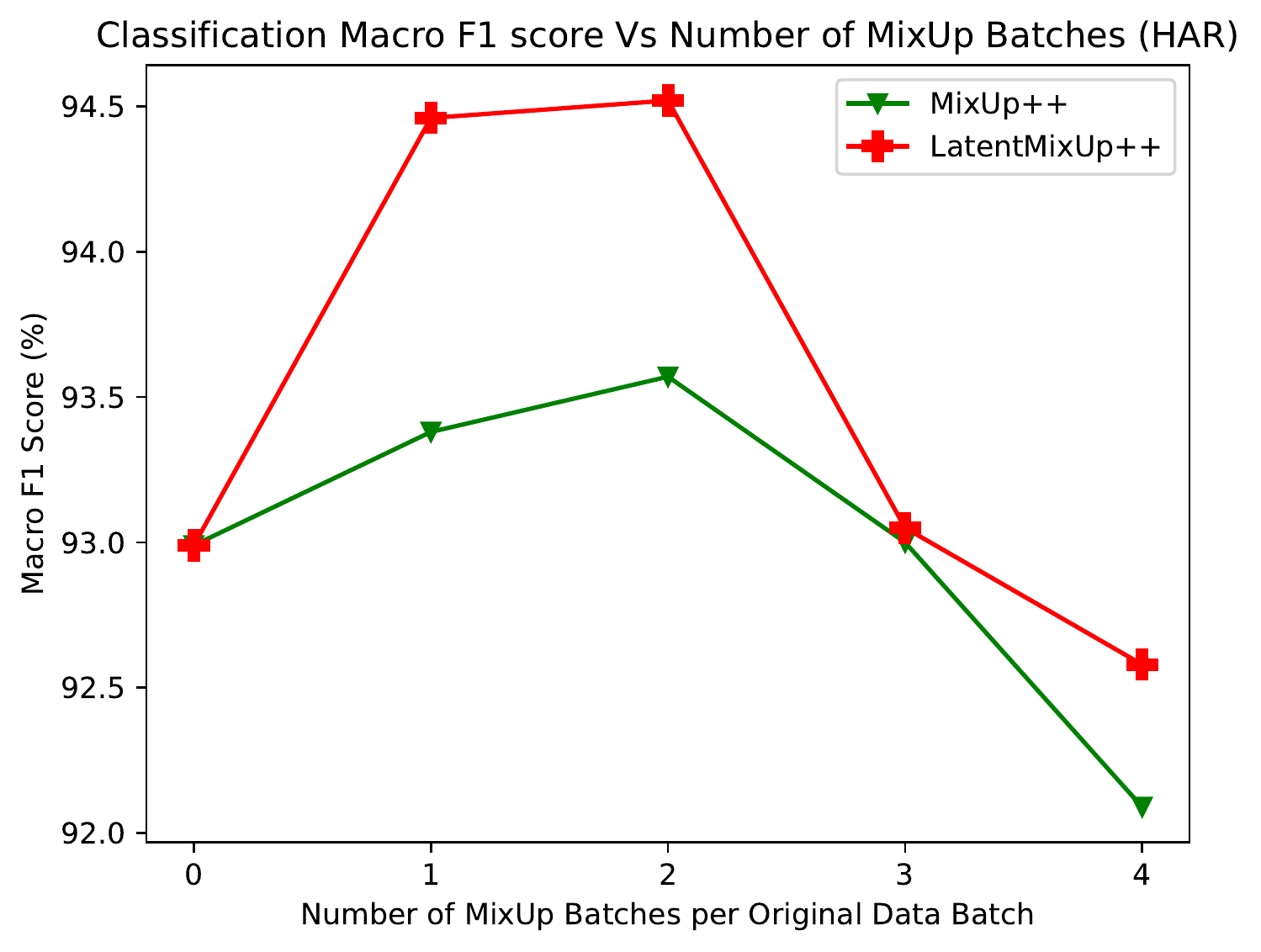}
\centering
\caption{ Classification Performance on HAR dataset for MixUp methods as a function of number of MixUp batches per original data batch.}
\centering
\label{fig:batches_ablation}
\end{figure}

\subsection{Ablation Study: Number of MixUp Batches}
In the previously shown results, we used MixUp methods with 2 batches of mixup data per original data batch. We present an ablation study in Figure~\ref{fig:batches_ablation}, on HAR dataset, as a function of the number of mixup data batches per original data batch. We observe a maximum when the number of batches is equal to two; performance drops subsequently as we increase the number of mixup batches. While this is empirical observation, we speculate that this is due to loss of decision boundary in the data manifold of the base model as the number of synthetic mixuped examples far exceeds the number of real examples.

\section{Conclusions}
\label{sec4:conc}
Time series are ubiquitous in nature, and commonly found in various domains. Labeling time series data is a challenging task as it needs domain expertise and more effort owing to the dynamic and temporal nature of the data. This often leads to situations with limited labeled data, with limited scope for getting more labeled data. Machine learning models are particularly data intensive. Data augmentation techniques have been used extensively in computer vision to overcome these issues. In this work, we extend MixUp to time series classification. MixUp is a commonly used simple technique in computer vision to interpolate input samples. We propose MixUp++ and LatentMixUp++ which use simple modifications over MixUp to perform interpolation in raw time series and classification model's latent space. We also propose extension of our methods in semi-supervised learning with pseudo labeling. Our results on two public datasets, indicate considerable improvement in performance for both low labeled data regime as well as high labeled data using LatentMixUp++.  

\bibliographystyle{plainnat}
\bibliography{iclr2021_conference}

\end{document}